\documentclass{article}

    \PassOptionsToPackage{numbers, compress}{natbib}

\usepackage[preprint]{neurips_data_2024}

\usepackage[utf8]{inputenc} %
\usepackage[T1]{fontenc}    %
\usepackage{hyperref}       %
\usepackage{tabularray}
\usepackage{url}            %
\usepackage{booktabs}       %
\usepackage{amsfonts}       %
\usepackage{nicefrac}       %
\usepackage{microtype}      %
\usepackage{xcolor}         %
\usepackage{amsmath}
\usepackage{multirow}
\usepackage{multicol}
\usepackage{graphicx}
\usepackage{makecell}

\title{Learning Traffic Crashes as Language: Datasets, Benchmarks, and What-if Causal Analyses}

\author{
  \textbf{Zhiwen Fan}$^{1\dagger}$,
  \textbf{Pu Wang}$^{2,3\dagger}$,
  \textbf{Yang Zhao}$^{3}$,
  \textbf{Yibo Zhao}$^{3}$,
  \textbf{Boris Ivanovic}$^{4}$,\\
  \textbf{Zhangyang Wang}$^{1}$,
  \textbf{Marco Pavone}$^{4,5}$,
  \textbf{Hao Frank Yang}$^{3*}$\\
  \normalsize{$\dagger$ Equal contribution}\quad  $*$ Corresponding author (haofrankyang@jhu.edu)\vspace{0.3cm}\\
  $^1$University of Texas at Austin\quad
  $^2$New York University  \quad \\
  $^3$Johns Hopkins University \quad
  $^4$NVIDIA Research \quad
  $^5$Stanford University \quad
}

\begin{document}

\maketitle

\begin{abstract}
The increasing rate of road accidents worldwide results not only in significant loss of life but also imposes billions financial burdens on societies. Current research in traffic crash frequency modeling and analysis has predominantly approached the problem as classification tasks, focusing mainly on learning-based classification or ensemble learning methods. These approaches often overlook the intricate relationships among the complex infrastructure, environmental, human and contextual factors related to traffic crashes and risky situations. In contrast, we initially propose a large-scale traffic crash language dataset, named \textbf{CrashEvent}, summarizing 19,340 real-world crash reports and incorporating infrastructure data, environmental and traffic textual and visual information in Washington State. Leveraging this rich dataset, we further formulate the crash event feature learning as a novel text reasoning problem and further fine-tune various large language models (LLMs) to predict detailed accident outcomes, such as crash types, severity and number of injuries, based on contextual and environmental factors. The proposed model, \textbf{CrashLLM}, distinguishes itself from existing solutions by leveraging the inherent text reasoning capabilities of LLMs to parse and learn from complex, unstructured data, thereby enabling a more nuanced analysis of contributing factors. Our experiments results shows that our LLM-based approach not only predicts the severity of accidents but also classifies different types of accidents and predicts injury outcomes, all with averaged F1 score boosted from 34.9\% to 53.8\%. Furthermore, CrashLLM can provide valuable insights for numerous open-world what-if situational-awareness traffic safety analyses with learned reasoning features, which existing models cannot offer. We make our benchmark, datasets, and model public available for further exploration.

\end{abstract}

\section{Introduction}\label{sec:intro}
\begin{figure}[ht]
  \centering
  \includegraphics[width=0.99\textwidth]{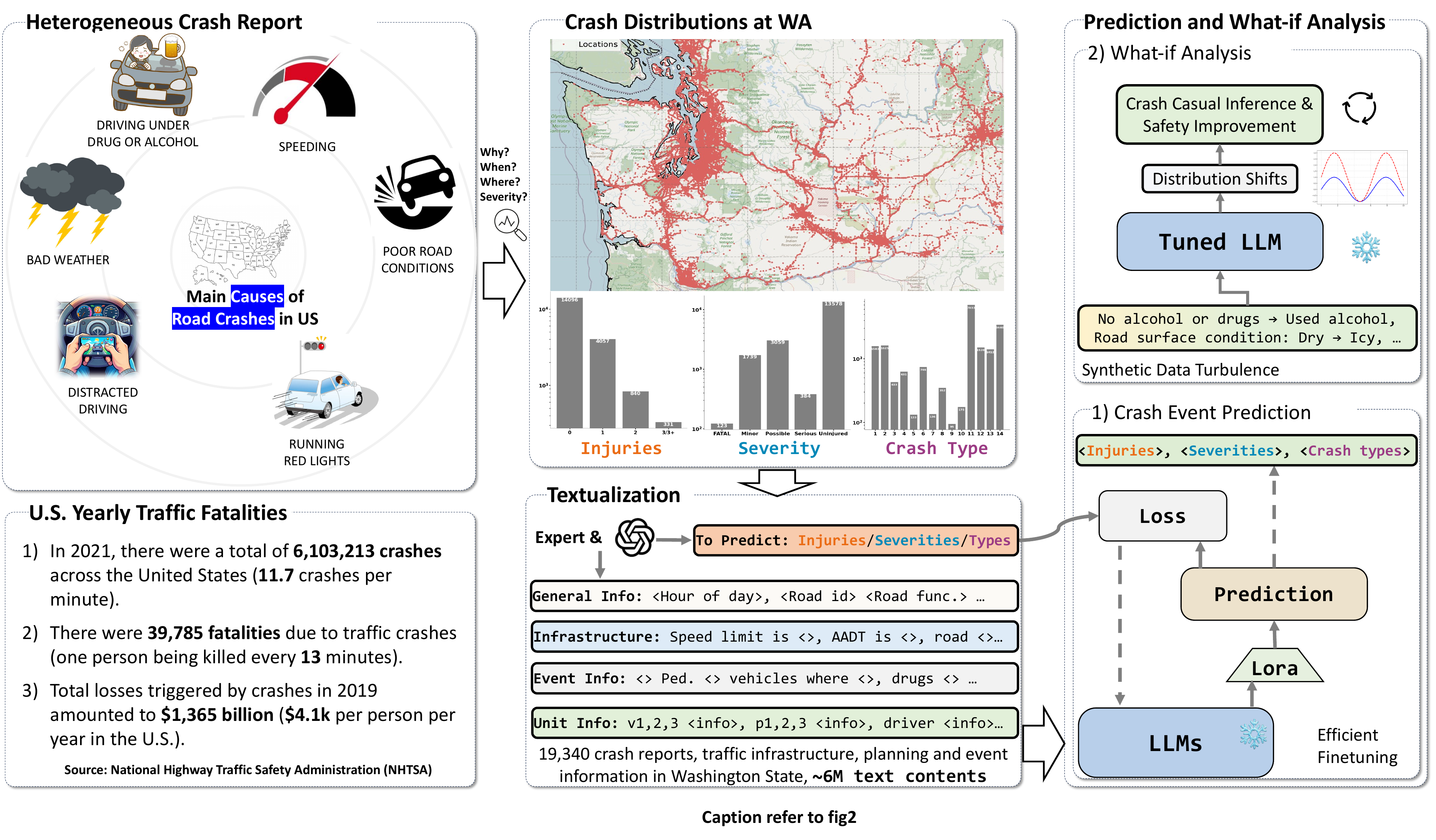}
  \vspace{-2mm}
\caption{\textbf{Overview of CrashEvent and CrashLLM.} Traffic crashes present a significant problem worldwide. We collected traffic crash records from Washington state and utilized textualization to reformulate these records. The tuned LLMs take this input, predict and analyze traffic crashes.}\label{fig:teaser}
  \vspace{-6mm}
\end{figure}

Road traffic crashes constitute a global public health crisis, resulting in substantial mortality, morbidity, and economic costs. In 2021, a total of 6,103,213 cases were reported in the United States. According to the National Highway Traffic Safety Administration (NHTSA) Fact 2022, on average, one crash occurs every five minutes, and 40.92\% of them result in injuries and long-term disabilities. Tragically, 39,785 of these crashes were fatal, resulting in one life lost every 11 minutes \footnote{https://crashstats.nhtsa.dot.gov/Api/Public/ViewPublication/813560}. In 2019, the total comprehensive loss for the USA was \$1.365 trillion, amounting to \$4,117 per citizen annually \cite{won2024alcohol}. These startling statistics underscore the urgent need for the research community to develop and implement effective interventions to save lives and reduce the economic burden. %
Typically, after a crash occurs, the details are summarized in a crash report by traffic agencies, utilizing figures, text and numerical formats to reconstruct the process. However, the causal factors of crashes are multifaceted, heterogeneous, and interconnected, encompassing a complex interplay of infrastructure design, human factors, environmental conditions, alcohol or drug use, vehicle-related factors, and other variables~\cite{wang2023analyzing}. This inherent complexity presents a longstanding challenge in analyzing these multimodal data and localizing the casual factors to learn from these tragedies.

Currently, existing researchers always use machine learning approaches~\cite{ahmed2023study} to formulate traffic accident analysis as classification tasks~\cite{SHARMA2020100127, xu2019statistical, MA2021106322}, summarizing and predicting crashes using a fixed number of features derived from heterogeneous crash reports. While these methods have provided valuable outputs, they typically oversimplify inputs into numerical categories and cannot offer accurate insights into event-level details. The process of discretizing text descriptions into handcrafted features (e.g., one-hot vectors, categorical levels) often fails to capture the complex inter- and intra-correlations among the diverse human, vehicle, behavior, regulation, environmental, and contextual factors present in textual crash records. Therefore, there is an obvious call for new approaches capable of learning from complex, unstructured crash text records to enable more accurate, reliable and useful prediction and reasoning analysis of crash contributing factors, thereby showing possibilities to improve the traffic safety effectively.

Recently, large language models (LLMs)~\cite{achiam2023gpt,touvron2023llama}, pretrained on extensive natural language data, have shown exceptional proficiency in contextual text reasoning capabilities with language information. The ability of LLMs to understand and generate human-like text suggests their potential for comprehending the complex and unstructured data found in crash reports. This understanding can facilitate the case level of analysis, what-if situational comparison, aiding in the identification of the hidden causes of accidents. However, to accurately interpret specific crash records, LLMs require fine-tuning due to the unique and nuanced nature of traffic data, which often includes heterogeneous information not adequately addressed by models trained on generic data.

To fill this gap, we introduce \textbf{CrashEvent} dataset and \textbf{CrashLLM}, the first event level traffic crash benchmark to LLMs, and investigates the LLMs' ability to forecast traffic crashes by reasoning textualized heterogeneous crash records.
CrashEvent comprises crash data from the whole Washington state of 2022 and defines three critical crash prediction tasks as text reasoning problem. Each crash instance in CrashEvent consists thorough descriptions in multiple aspects (see Figure~\ref{fig:teaser}), allowing the integration of original and rich crash reports without the loss of original textual information.
CrashEvent is curated with human-machine cooperative approach contains two-phase process: 1) Crash Textual Recategorization and Organization, involving human experts to formulate structured and meaningful paragraph-wise forms, including general information, infrastructure, even information and crash unit information.
2) Machine-guided Crash Report Generation. In this phase, we ask ChatGPT to transform and fill the original unstructured text into crash template.
Our two-phase procedure ensures the data generation process is highly efficient and requires minimal human labelling efforts, to generate rich and lossless crash contexts.
Then, we conducted experiments using both standard efficient fine-tuning with LoRA~\cite{hu2021lora} of LLM and representation ML methods to perform event-level crash predictions. Our finding indicate that by formulating the traffic crash prediction as text reasoning problem, CrashLLM can significantly outperform all traditional methods to leverage the uncompressed data format.
For proactively mitigate risks and enhance overall traffic safety, we perform conditional what-if analysis by visualizing the distribution shift by synthesizing the test data cases with perturbing specific attributes. The analysis allow the inditification of improving safety predictions and equip first responders with timely, event-specific insights, ultimately reducing the chances of crashes happening. We summarize the contributions as follows:
\begin{enumerate}
\item We introduce the CrashEvent dataset, comprising 19,340 crash records from 242 cities and 1,973 road segments in Washington State during 2022, totaling approximately 6.32 million words. Each crash event record includes 50 attributes describing the infrastructure, event, environment, and textual descriptions of the vehicles and pedestrians involved.
\item We introduce CrashLLM, the first event-level traffic crash prediction framework. We demonstrate the effectiveness of fine-tuning CrashLLM for traffic crash prediction tasks by formulating crash prediction as text-based reasoning analysis.
\item Experimental results show that CrashLLM achieves an average F1 score 18.89\% higher on three tasks (injury, severity, and accident type prediction) compared to existing machine learning models. We further conduct a what-if situational-aware analysis by synthesizing test data to explore hypothetical scenarios and assess their potential impact on traffic safety outcomes.
\end{enumerate}
\section{Related Works}
\vspace{-3mm}
\paragraph{Existing Approaches to Learning Traffic Crash Forecasting} 
Multiple traditional methods have been employed to analyze traffic, focusing primarily on predicting injury severity using machine learning techniques. Some studies frame the problem as binary classification \cite{SHARMA2020100127, xu2019statistical, MA2021106322}. For instance, Assi (2020) developed a hybrid system using Principal Component Analysis (PCA) with Multi-Layer Perceptron (MLP) and Support Vector Machine (SVM) to predict traffic crash severity, distinguishing between slight injury and serious/fatal \cite{assi2020traffic}. Other studies consider it as multiclass classification for different severity levels \cite{OSMAN2016261, NAIK201657, JEONG2018250, chen2020analysis, jamal2021injury}. For example, Satter et al. (2023) predicted injury levels using vanilla MLP with embedding layers, and TabNet \cite{sattar2023transparent}. Most studies utilize text-based traffic crash reports \cite{s20041107}, but machine learning methods require transforming text into numerical representations, which may cause information loss and impact prediction accuracy. Additionally, these models only learn the distribution of the training data and lack true understanding of underlying causality, limiting their ability to generalize and predict accurately when distributions change.

\vspace{-3mm}
\paragraph{Related Causality and Factors Influencing Traffic Crashes} 
In some of these studies, researchers have delved further into investigating the relationship between risk factors and crash severity. Various machine learning methods have been widely used to solve the crash injury severity classification problem and analyze the contributing factors related to injury severity. These methods include Support Vector Machine (SVM) \cite{zhang2024machine, YU201450}, Logistic Regression (LR) \cite{8717393, CANDEFJORD2021101124}, AdaBoost \cite{8717393}, Bayesian network \cite{CHEN201576, JEONG2018250, DEONA2011402, CHEN201695}, K-means Clustering and Latent Class Clustering \cite{iranitalab2017comparison, sun2022exploring}, and SHapley Additive exPlanations (SHAP)~\cite{lundberg2017unified}. 
However, the real-world crash data distribution is significantly imbalanced with incomplete and heterogeneous records, which can introduce bias and inaccuracies in the model's learning process.

\vspace{-4mm}
\paragraph{Human-Like Reasoning in Large Language Models}
Recent advancements in large language models (LLMs), such as GPT-4 \cite{achiam2023gpt} and LlaMA 2 \cite{touvron2023llama}, have expanded the scope of artificial intelligence from traditional predictive analytics to emulating complex human-like interactions in various systems~\cite{gao2023s}. One notable feature within LLMs is in-context learning (ICL), where the model performs tasks based on input-output examples without parameter adjustments. Additionally, knowledge extraction from locally deployable LLMs through fine-tuning, particularly using Parameter-Efficient Fine-Tuning (PEFT) techniques \cite{hu2021lora,dettmers2023qlora}, offers valuable insights. We will adopt the PEFT approach \cite{hu2021lora} for fine-tuning CrashLLM.
\section{Transforming Non-Numeric Crash Events into Textual Format}\label{sec:datasets}
\begin{figure}[ht]
  \centering
  \includegraphics[width=1\textwidth]{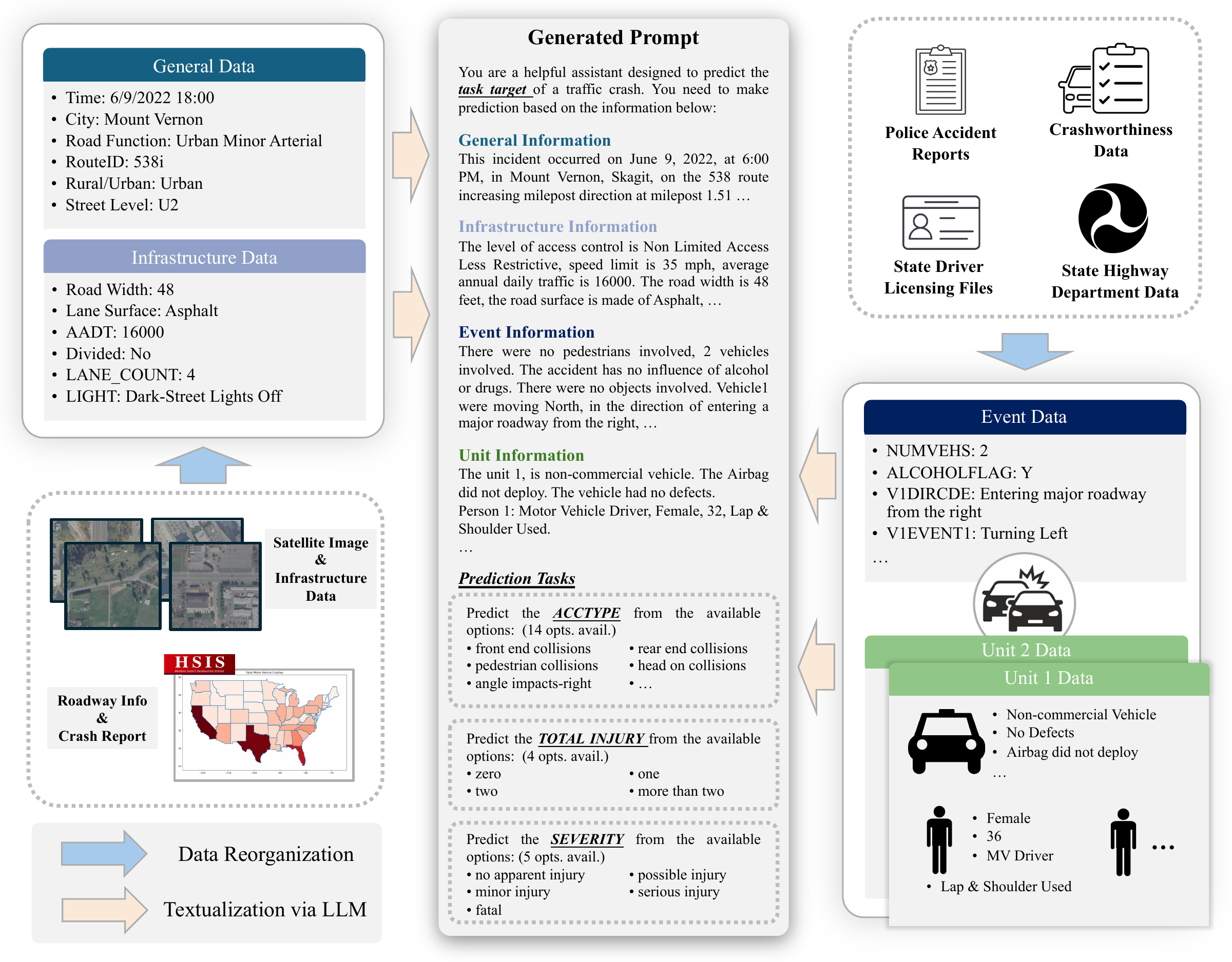}
  \vspace{-6mm}
\caption{\textbf{Illustration of data pre-processing and text Prompt Design.} We illustrate the process from raw data from HSIS, satellite images, crash reports to textualized prompts in CrashEvent dataset. The original data are reorganized into general data and infrastructure data. Additionally, police accident reports, crashworthiness data, driver license data, and state highway department data are reorganized based on events into event data and unit-based data. These \textbf{four} types of data are textualized into approximately 300-word paragraphs aided by ChatGPT. The blue arrow in the figure represents data reorganization, while the beige arrow represents the textualization via LLM.}
\label{fig:data_processing}
\end{figure}
Describing an accident involves complex and diverse information, including the environment of the accident location, details of the vehicles involved, and a thorough description of the accident process. Providing such detailed information is essential for understanding the accident and analyzing its causes~\cite{montella2013crash}. To obtain a comprehensive accident dataset, we gathered data from a variety of sources and formats, then reorganized and processed it into a unified dataset for each accident. Finally, we categorized the dataset into four types of information, which is general information, infrastructure information, event information and unit information, and convert them into approximately 300-word text descriptions to serve as the final input dataset. Our datasets with license description can be accessed at ~\url{https://crashllm.github.io/}.

\vspace{-4mm}
\paragraph{Introduction to Raw Data Sources and Types}
Our dataset includes crash data from Washington State in 2022, in total of 19,340 records after filtering significantly incomplete data. The primary sources of our dataset include Highway Safety Information System (HSIS) crash data\footnote{\url{https://highways.dot.gov/}}, satellite image data, police accident reports, crashworthiness data and state driver licensing data. The HSIS data consists of two components: one describes the physical layout of roads and the associated traffic characteristics within Washington State, and the other consisting of crash reports that provide a general description of accidents. Satellite images are obtained based on the location infrastructure and planning information, i.e., intersection alignments, neighborhood types, in the crash reports. By processing these satellite images, we can obtain more detailed descriptions of the road infrastructure. Police accident reports contain detailed descriptions of accidents, with their data elements standardized into a common format. Crashworthiness data includes information collected from crash sites and data related to the vehicles involved in the accidents. State driver licensing data includes basic information about the individuals involved in accidents, such as age and gender. Figure~\ref{fig:data_processing} uses an example of crash data to illustrate the process of generating a prompt from the raw data.

\vspace{-3mm}
\paragraph{Feature Engineering and Textual Organization of Crash Data}
For each accident, we associated the crash report with the involved vehicles and individuals using the crash report number, thus obtaining descriptions of the accident and the persons involved. The route ID and milepost were used to identify the specific road segment where the accident occurred, allowing us to gather related roadway data in existing database. Additionally, to supplement the infrastructure and environmental information, we obtained satellite images based on GPS coordinates and used VLLM~\cite{achiam2023gpt} to supplement environmental information. 
Due to the diverse sources of data, we performed feature cleaning by deleting or merging duplicate features. Given the substantial amount of textual descriptions in the data, we employed an AI-human collaborative approach for dimensionality reduction on certain features. This process generalized the data and reduced redundancy. Based on the five W's (where, when, what, who, why) of crash reports \cite{imprialou2019crash}, we categorized the data into four types of information (general information, infrastructure information, event information and unit information). Finally, for providing logically coherent and continuous textual data which is amenable to LLM learning, we transformed each category of data into text format using an AI-human cooperative prompt design. The mentioned work provides a comprehensive textual database that describes the accident environment, process, and entities involved, forming a solid foundation for LLM learning on accident data. Finally, after filtering out datapoints with significant missing information, the \textbf{CrashEvent dataset} mergers the complementary information from multimodal data sources and contains 19,800 crash records with approximately 6.32 million words for further use.

\begin{table}[h]
	\normalsize
	\centering
	\begin{center}
\caption{\textbf{Crash Prediction Explanation}. We show how we formulate the tasks, including \textcolor{blue}{Severity} and \textcolor{purple}{Accident Type}, with detailed definitions of each label. We omit the task of \textcolor{orange}{Injury} in the table.
\textbf{No.} represents the label ID, and \textbf{Abbr.} indicates the abbreviation.}
		\label{tab:definitions:severity_and_acctype}
		\setlength{\tabcolsep}{6pt} %
		\renewcommand{\arraystretch}{1.3} %
		\resizebox{0.89\textwidth}{!}{
			\begin{tabular}{p{1.5cm}|c|c|c|p{9.5cm}}
				\toprule[1.5pt]
				\textbf{Variable}                          & \textbf{No.} & \textbf{Values}            & \textbf{Abbr.} &\makecell{ \textbf{Definitions}}                                                                     \\
				\hline
				\multirow{5}{*}[-3ex]{\makecell{\textcolor{blue}{Severity}\\(S)}}        & 1            & No Apparent Injury         & O              & No visible injuries reported at the scene.                                                                \\
				                                           & 2            & Possible Injury            & C              & Any injury reported to the officer or claimed by the individual.                                          \\
				                                           & 3            & Minor Injury               & B              & Any injury other than fatal or disabling at the scene.                                                    \\
				                                           & 4            & Serious Injury             & A              & Any injury that prevents an individual from walking, driving, or continuing their normal activities.      \\
				                                           & 5            & Fatal                      & K              & Any injury that directly results in the death of a living person within 30 days of a motor vehicle crash. \\
				\hline
				\multirow{14}{*}[-1ex]{\makecell{\textcolor{purple}{Accident}\\\textcolor{purple}{Type} (AT)}} & 1            & Single Vehicle With Object & SVO            & Collision involving a single vehicle and a stationary object.                                             \\
				                                           & 2            & Angle Impacts Right        & AIR            & Vehicles collide at an angle, impacting on the right side.                                                \\
				                                           & 3            & Other                      & Oth            & Any other types of accidents not classified in specific categories.                                       \\
				                                           & 4            & Sidewipes Left             & SL             & Vehicles sideswipe each other on the left side.                                                           \\
				                                           & 5            & Front End Collision        & FEC            & Collisions where the front ends of vehicles impact each other.                                            \\
				                                           & 6            & Rear End Collision         & REC            & Collisions where one vehicle impacts the rear of another.                                                 \\
				                                           & 7            & Overturn                   & OT             & Accidents where a vehicle overturns.                                                                      \\
				                                           & 8            & Animal Collision           & AC             & Collisions involving animals.                                                                             \\
				                                           & 9            & Pedestrian Collision       & PC             & Accidents where a vehicle collides with a pedestrian.                                                     \\
				                                           & 10           & Sidewipes Right            & SR             & Vehicles sideswipe each other on the right side.                                                          \\
				                                           & 11           & Pedal Cyclist Collision    & PCC            & Collisions involving cyclists.                                                                            \\
				                                           & 12           & Head On Collision          & HOC            & Head-on collisions between vehicles.                                                                      \\
				                                           & 13           & Off Road                   & OR             & Accidents involving vehicles going off the road.                                                          \\
				                                           & 14           & Angle Impact Left          & AIL            & Vehicles collide at an angle, impacting on the left side.                                                 \\
				\bottomrule[1.5pt]
			\end{tabular}
   \vspace{-4mm}
		}
	\end{center}
\end{table}   \vspace{-3mm}

\paragraph{Defining Inputs and Outputs}
In the context of a traffic accident, the outcome and severity are of primary concern. Numerous studies focus on predicting and analyzing the types of accidents, their severity, or the number of injuries involved \cite{abdel2005analysis, IRANITALAB201727, SAVOLAINEN20111666}. To effectively measure these aspects, we selected \textbf{three} variables from the accident reports to describe the outcome of the accident: 
the number of people \textcolor{orange}{injured} \((\mathcal{I}_t)\), which is more balanced compared to using the raw number of injuries, the \textcolor{blue}{severity} of the accident on the KABCO scale~\footnote{https://highways.dot.gov/media/20141} \((\text{S})\), which is commonly utilized in police-recorded accident data \cite{mujalli2013injury} and the \textcolor{purple}{accident type} \((\text{AT})\). We utilize these three variables to describe the crash result \((\text{CR}_i)\), where \(i\) denotes the unique identifier caseid. The accident outcome can be presented in the following format: $\text{CR}_i=\text{AT}_S^{\mathcal{I}_{t}}$. The function \(\mathcal{I}_t\)  describes the number of people injured in an accident as follows: zero if t \(=0\), one if t \(=1\), two if t \(=2\), and more than two if t \(\geq 3\), where t represents the number of people injured. The values for S and AT are provide in the Table~\ref{tab:definitions:severity_and_acctype}.

For the model's input, four segments of textual information are contained, as shown in Figure~\ref{fig:data_processing}. Each paragraph consists of approximately 100 words. Together, they provide a comprehensive and detailed description of the accident. The content of each paragraph is outlined below: \textbf{1) General Information:} this includes specifics about the time and location of the accident, as well as the type of road where it occurred. \textbf{2) Infrastructure Information:} this covers descriptions of the road infrastructure, including both static elements like the number of lanes and speed limits, and dynamic features such as work zone indicators, lighting, and road surface conditions at the time of the accident. \textbf{3) Event Information:} this segment provides a detailed account of the accident process and the contributing factors identified in the records. \textbf{4) Unit Information:} this involves details about the vehicles and individuals involved in the crash.

\vspace{-3mm}
\section{Adapting Language Models for Text-Based Crash Reasoning Analysis}\label{sec:method}

We adapt LLaMa-2~\cite{touvron2023llama} to crash prediction tasks to enhance the LLMs' capabilities in interpreting crash data, identifying critical factors, and conducting causality analysis to offer insights for crash prevention.

\vspace{-3mm}
\paragraph{Construct Training Data for LLMs}
In the training of large language models (LLMs), a single input consists of three components: the system prompt, the user prompt, and the target prompt. Details regarding the system and user prompts are presented in Section~\ref{sec:datasets}. The target prompt is formulated using the template: "The answer is: <PREDICTION>", where <PREDICTION> represents the ground truth. These components are structured as follows: "System: <system prompt>, User: <user prompt>, Assistant: <target prompt>". We use LLaMA-2's tokenizer to segment the text inputs into tokens.

\vspace{-3mm}
\paragraph{Additional Special Tokens for Classification}
To adapt the LLM as a crash classifier, additional tokens have been incorporated into the tokenizer's vocabulary. Specifically, for predicting the total number of injuries, four special tokens have been introduced: [<ZERO>, <ONE>, <TWO>, <THREE OR MORE>], corresponding to zero, one, two, and three or more injured individuals, respectively, in the crash event. This approach has also been applied to the predictions of severity and accident type (see supplementary materials for details). The parameters of the input and output embedding layers are set as trainable, enabling the model to align the representations of these special tokens with the existing embedding space.
\vspace{-3mm}
\paragraph{Supervised Finetuning}
During the fine-tuning phase, the traffic forecasting task is framed as a next-token generation task. This process can be described as:
\vspace{-1mm}
\begin{equation}
   p_{\theta}(T_i)=\prod_{j=1}^{|T_i|}{p_{\theta}(t_j^{(i)} \vert t_{1}^{(i)}, \cdots , t_{j-1}^{(i)}}),\vspace{-1mm}
\label{eq: autoregressive}
\end{equation}
where $T_i$ is the $i$-th item in the training data, $p_{\theta}$ is the LLM model, $t_j^{(i)}$ denotes the $j$-th token in $T_i$. By maximizing the likelihood $ p_{\theta}(T)=\prod_{i=1}^{N}{p_{\theta}(T_i)}$, the LLM's parameters are learned. Both the system prompt and the user prompt are masked for loss computation during training. Through this process, the model learns to make prediction for a traffic crash accident. In our experiments, we utilize LoRA~\cite{hu2021lora} to fine-tune LLaMA-2 models. All the models are loaded in 4-bit. We use AdamW~\cite{loshchilov2017decoupled} as optimizer and train the models on Nvidia A100 80GB GPU with DeepSeed~\footnote{https://www.deepspeed.ai/}.

\vspace{-2mm}
\section{Experiments}\label{sec:exp}
\vspace{-2mm}
We conduct experiments to study the effectiveness of the trained model, we consider two research evaluation settings: (1) what is the performance of CrashLLM as a traffic crash predictor? (2) How was the conditional analysis ability to see how LLMs have mastered conditional and modal reasoning? 

\vspace{-3mm}
\paragraph{CrashEvent Dataset Split and Evaluation Metrics}

We split the collected data into two parts: data from January, June, and December are used as the testing set (842 data points), with additional resampling based on the total number of injuries to create a uniformly distributed evaluation subset. The remaining data (15,014 data points) are used as the training set. All experiments were configured to predict tasks including \textcolor{orange}{Total Injuries}, \textcolor{blue}{Crash Severities}, and \textcolor{purple}{Accident Types}. 
In evaluating the model performance as a classification task, we employ accuracy, precision, recall, and F1 score as metrics.

\vspace{-3mm}
\paragraph{Adopted Baselines}
We follow the recent literature~\cite{ahmed2023study} and also adopt Random forest (RF)~\cite{randomforest}, Decision Trees (DT)~\cite{quinlan1986induction}, Adaptive boosting (AdaBoost)~\cite{freund1999short}, Bayesian Network(BN)~\cite{deleu2022bayesian}, LogisticRegression (LR)~\cite{cox1958regression}, and Categorical boosting (CatBoost)~\cite{prokhorenkova2018catboost} as compared baselines.
\begin{table}[h]
    \small
    \begin{center}
\caption{\textbf{Performance Comparison} of the three Crash Prediction tasks. We present quality metrics along with model rankings by averaging the column-wise rank.}
    \label{tab:performance:AccLLM}
    \setlength{\tabcolsep}{4pt} %
    \renewcommand{\arraystretch}{1.1} %
    \resizebox{0.99\textwidth}{!}{
    \begin{tabular}{c|c|c|c|c|c}
    \toprule[1.5pt]
      \multirow{3}{*}{\textbf{Model}} & \multicolumn{4}{c|}{\textbf{Evaluation Metric (Model Rank)}}  & \multirow{3}{*}{\textbf{Rank}}\\
     \cline{2-5}
      & \raisebox{-1ex}{\textbf{Accuracy $\uparrow$}} & \raisebox{-1ex}{\textbf{Precision $\uparrow$}} & \raisebox{-1ex}{\textbf{Recall $\uparrow$}} & \raisebox{-1ex}{\textbf{F1-score $\uparrow$}} \\
     & \text{\small \textcolor{orange}{Injury}/\textcolor{blue}{Severity}/\textcolor{purple}{Type}} & \text{\small \textcolor{orange}{Injury}/\textcolor{blue}{Severity}/\textcolor{purple}{Type}} & \text{\small \textcolor{orange}{Injury}/\textcolor{blue}{Severity}/\textcolor{purple}{Type}} & \text{\small \textcolor{orange}{Injury}/\textcolor{blue}{Severity}/\textcolor{purple}{Type}}
      \\
      \hline

     RandomForest~\cite{breiman2001random} & 0.353 / 0.339 / 0.384 & 0.124 / 0.115 / 0.543 & 0.353 / 0.339 / 0.384 & 0.184 / 0.171 / 0.395 & 6.58 (9)\\
AdaBoost~\cite{freund1999short} & 0.353 / 0.339 / 0.579 & 0.124 / 0.115 / 0.383 & 0.353 / 0.339 / 0.579 & 0.184 / 0.171 / 0.447 & 6.33 (8)\\
CatBoost~\cite{prokhorenkova2018catboost} & 0.353 / 0.339 / 0.702 & 0.124 / 0.115 / 0.664 & 0.353 / 0.339 / 0.702 & 0.184 / 0.171 / 0.667 & 5.08 (6)\\
Bayesian Network~\cite{pearl1988probabilistic} & 0.394 / 0.341 / 0.653 & \textbf{0.485} / 0.306 / 0.563 & 0.394 / 0.341 / 0.653 & 0.287 / 0.181 / 0.578 & 4.67 (4)\\
DecisionTree~\cite{quinlan1986induction}& 0.353 / 0.347 / 0.677 & 0.124 / 0.207 / 0.631 & 0.353 / 0.347 / 0.677 & 0.184 / 0.190 / 0.640 & 4.75 (5)\\
LogisticRegression~\cite{cox1958regression}& 0.353 / 0.339 / 0.566 & 0.124 / 0.115 / 0.471 & 0.353 / 0.339 / 0.566 & 0.184 / 0.171 / 0.457 & 6.25 (7)\\
   \cline{1-5}
     Llama-7B          & 0.399 / 0.382 / 0.740 & 0.404 / 0.411 / 0.771 & 0.399 / 0.382 / 0.740 & 0.401 / 0.379 / 0.744  &2.92 (3)\\
     Llama-13B           & 0.439 / 0.393 / \textbf{0.748} & 0.431 / 0.375 / 0.767 & 0.439 / 0.393 / \textbf{0.748} & 0.427 / 0.353 / 0.755 &2.08 (2)\\
     Llama-70B           & \textbf{0.447} / \textbf{0.436} / 0.747 & 0.451 / \textbf{0.446} / \textbf{0.775} & \textbf{0.447} / \textbf{0.436} / 0.747 & \textbf{0.445} / \textbf{0.411} / \textbf{0.757} &\textbf{1.25} (1)\\

    \bottomrule[1.5pt]
    \end{tabular}
    }
    \end{center}
\end{table}
\begin{figure}[ht]
  \centering
  \includegraphics[width=0.9\textwidth]{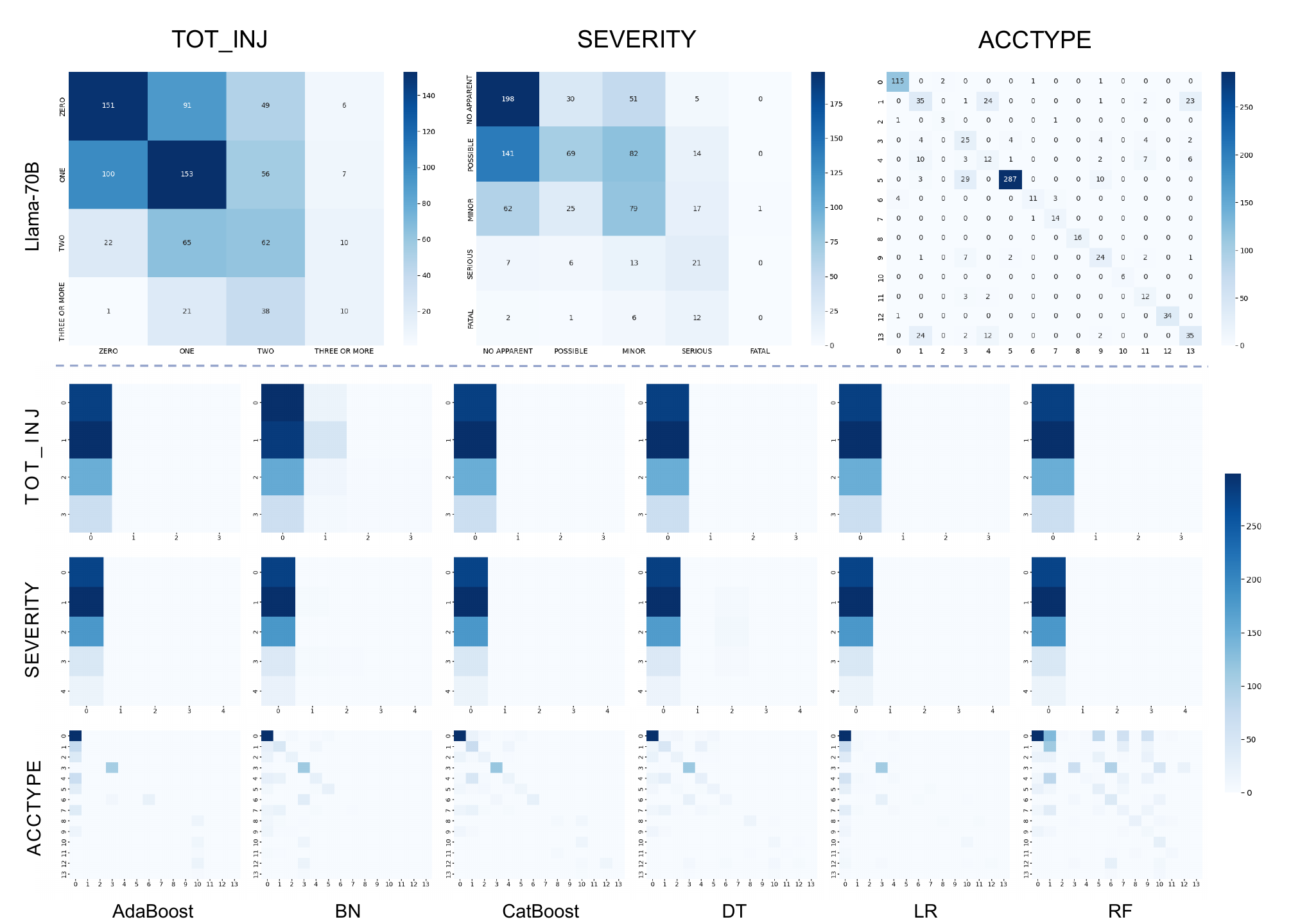}
  \vspace{-2mm}
\caption{\textbf{Summary of Model Confusion Metrics.} We display the confusion matrix for our top-performing model, LLama-70b, compared to all baseline models. Baseline models tend to predict the category with the highest number of instances (e.g., zero injury, no apparent injury). Our CrashLLM, on the other hand, is not constrained to predict a specific class, validating that the improved accuracies arise from enhanced reasoning capabilities.}\label{fig:cm} \vspace{-5mm}
\end{figure}
\vspace{-6mm}

\paragraph{Comparisons with SoTA on Crash Prediction.}
The quantitative comparisons between CrashLLM and other established machine learning models are shown in Table~\ref{tab:performance:AccLLM}. In this table, "inj" represents the task of predicting the total number of people injured, "sev" represents the task of predicting the severity, and "acc" represents the accident type task. Figure~\ref{fig:cm} presents the confusion matrix results for the baseline models. Here, we formulate traffic crash prediction as a text classification problem, categorizing crashes' injuries into four categories, severity into five categories, and crash type into fourteen categories. A comprehensive examination of existing crash frequency and prediction models shows that none reliably provide useful outcomes at the event level. Most of these models prioritize fitting distributions dominated by head categories, rather than learning distinct crash features. While some results, such as the BN on injury prediction, appear promising, the confusion matrix demonstrates its tendency to predict only one head categories. Promisingly, we observe that CrashLLM outperforms all other baselines in the averaged metrics, with the 70B model performing the best on average for all adopted tasks. This validates CrashLLM's robust capability in following instructions to predict traffic crash properties. A reliable forecasting model that performs accurate reasoning based on crash records is crucial to prevent significant prediction errors. To further evaluate the reliability of all adopted models, we visualize the confusion matrices in Figure~\ref{fig:cm}. We can observe that traditional classification ML models tend to predict the dominant categories (e.g., zero injury under \textcolor{orange}{Total Injury}, no apparent injury \textcolor{blue}{Crash Severity}), which are mostly in the first column. In contrast, CrashLLM utilizes its text reasoning capacity to predict traffic crashes by leveraging complex, heterogeneous text data. This suggests that CrashLLM can offer valuable insights for making informed operational decisions through detailed model analysis.

\vspace{-3mm}
\paragraph{What-if Situational Analysis for Informed Transportation.}
Transportation is a complex system that directly interacts with human beings. For traffic agencies, changing safety policies and conducting analyses typically require considering many scenarios. It is impossible to collect data for and encode all of these scenarios into traditional machine learning models, as discussed in previous sections. For example, many DoTs previously tried to investigate how an increase in people driving under the influence of drugs and alcohol would change the severity of traffic crashes \cite{won2024alcohol,fell2014effects}, but can not be done without more useful samples. Similar questions are raised by numerous agencies and policymakers, to name just a few: What if the weather conditions changed from sunny to snowy? How would this impact traffic safety? What if there are work zones on the road? How would this affect the types of crashes and the severity of injuries? What if the road conditions were icy? How could we estimate the impact on the distribution of traffic crashes? These kinds of "what-if" situational-awareness questions are frequently posed, but no model could answer them until the introduction of CrashLLMs. Compared to traditional classification-based ML models, one of the biggest advantages of LLMs is their human-like text reasoning capabilities. Given their extensive vocabulary and ability to infer real-world human logic, LLMs offer a unique advantage which can be used to explore hypothetical scenarios and assess their potential impact on traffic safety outcomes. And we prove the fine-tuned CrashLLM with this capabilities and can provide valuable conditional comparisons and be used as human decision references. In this study, we focused on three key factors known to impact traffic crashes \cite{ditcharoen2018road,osman2019impacts}: alcohol and drug use, adverse weather conditions (specifically icy roads), and the presence of work zones. To investigate their effects, we synthesized three scenarios in the test set by incrementally converting a portion of the original conditions (\textbf{non-alcohol, dry roads, non-work zone}) into their corresponding adverse conditions (\textbf{alcohol, icy roads, work zone}). By perturbing the original testing sets at the rates of 100\% (double the impact cases), 200\% (triple), and finally all cases, simulating increasing levels of these risk factors, compared with the original distribution. 

\begin{figure}[h]
  \centering
  \includegraphics[width=0.99\textwidth]{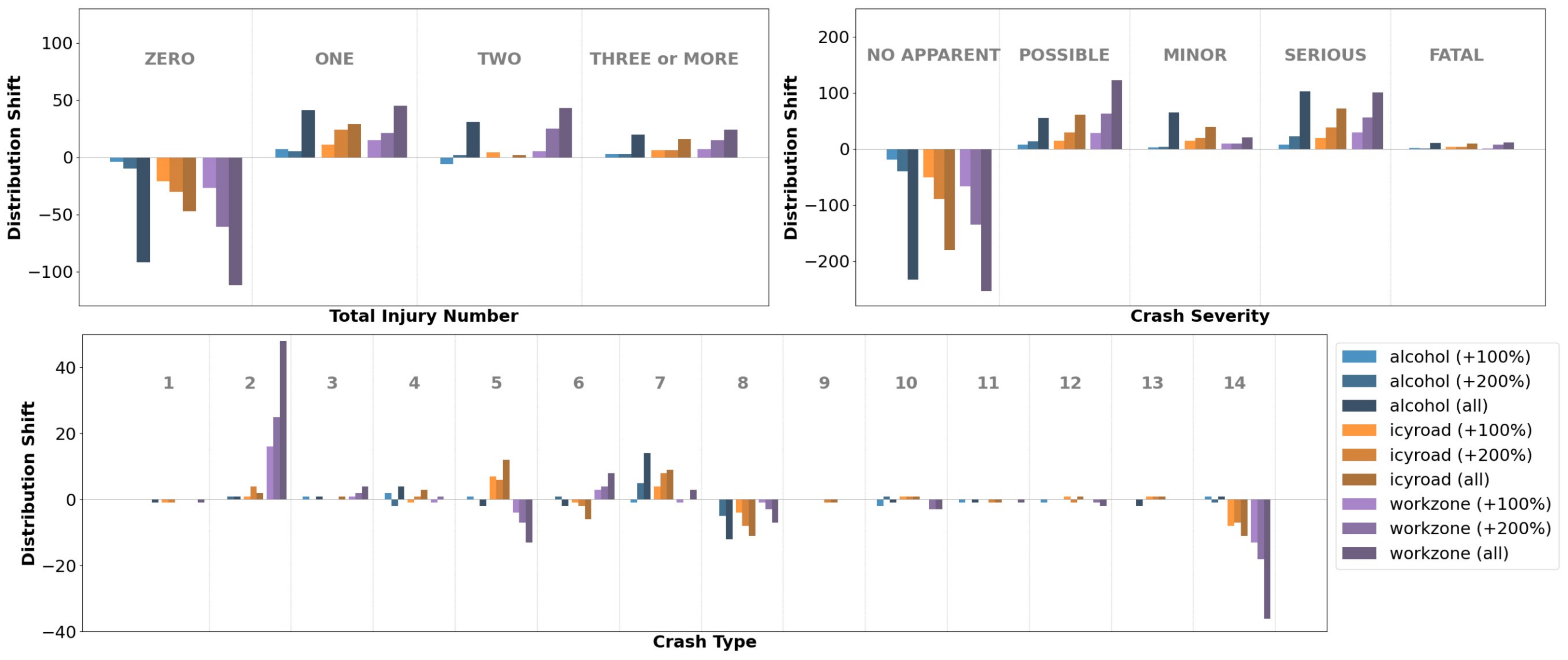}
  \vspace{-2mm}
\caption{\textbf{What-if Situational Analysis.} In the three figures, all the number of y-axis are case changes without/with perturbations. The x-axis, zero, represent there are no changes after perturbing the data. The total crash case number in the testing set remains unchanged.}\label{fig:what_if}
\vspace{-4mm}
\end{figure}

The results, illustrated in Figure~\ref{fig:what_if}, demonstrate a clear distribution shift in crash outcomes. Even without additional fine-tuning, our CrashLLM model effectively captures the influence of these factors on crash likelihood and severity. Among the three factors, work zone existence has the most significant impact on crash distributions. Doubling work-zone-related crash test cases triggers a 21\% increase in crashes with 3 or more injuries, leads to a 42\% higher percentage of serious crashes, and results in a 20\% higher rate of right-side angle impacts due to necessary lane changes. Doubling the number of alcohol-involved crashes leads to a 10\% increase in serious injury crashes and a 200\% higher rate of fatal crashes. Alcohol use also triggers a 16\% increase in crashes involving pedestrians and cyclists, and a 7\% increase in rollovers. Icy road conditions have the most substantial impact on changes in serious injury crashes. Doubling these cases leads to a 27\% increase in crashes causing serious injury, with a 30\% higher rate of rollovers and a 12\% increase in angle impacts. More what-if analysis and casual findings can be found in the supplementary materials. These findings and what-if situational comparisons highlight the model's ability to leverage existing knowledge and generate insightful predictions even when faced with data limitations.

\vspace{-3mm}
\section{Conclusion}
\vspace{-3mm}
In this study, we introduced a new traffic crash prediction dataset, CrashEvent, by textulizing heterogeneous crash records into a language-based representation and developed a traffic crash prediction framework, CrashLLM, that leverages the advanced capabilities of large language models (LLMs) to analyze and predict traffic crash incidents. We demonstrated that CrashLLM outperforms established machine learning models across multiple tasks. By utilizing complex, heterogeneous data through text reasoning, it allows for a deeper understanding of the underlying factors contributing to traffic crashes. This capability is crucial for developing informed operational strategies and making data-driven decisions for enhancing city-level infrastructures.
\vspace{-3mm}
\section{Limitations, Future Works and Societal Impacts}\label{sec:limitation_and_impact}
\vspace{-3mm}
Despite the successes of our framework in demonstrating promising performance in traffic crash prediction, CrashLLM requires separate training for each adopted task. A potential solution would be to incorporate new versions of LLMs with specifically designed prompts to denote different prediction tasks within a unified model.
Our research enables more accurate event-level crash predictions. This technology is advantageous for crash forecasting and enhances overall traffic roadway safety.

\newpage
\section*{Technical Appendices}
\begingroup
\fontsize{9pt}{11pt}\selectfont
This technical appendices provides more details which are not included in the main paper due to space limitations.
We have included few prompt examples, 
detailed description of added special tokens, explaination about what-if analysis, and the generation process of satellite images. Our organized CrashEvent datasets can be accessed through the \href{link}{crashllm.github.io}. 

\paragraph{Prompt Examples.}
In our research, we utilize textualized prompts to facilitate model understanding across different tasks. We illustrate this with examples from three distinct tasks: traffic injury prediction, crash severity classification, and accident type estimation. We show five different prompts below, showcasing how we structure the input data into prompts that the model can process effectively.

\paragraph{Additional Special Tokens for Classification}
As shown in the main draft, for predicting the \textcolor{orange}{Total Injuries}, we have introduced four special tokens: [<ZERO>, <ONE>, <TWO>, <THREE OR MORE>], representing zero, one, two, and three or more injured individuals in a crash event, respectively. Similarly, for predicting the \textcolor{blue}{Crash Severities}, we use five additional tokens: [<NO APPARENT INJURY>, <POSSIBLE INJURY>, <MINOR INJURY>, <SERIOUS INJURY>, <FATAL>], corresponding to different levels of severity. For the task of identifying \textcolor{purple}{Accident Types}, we utilize 14 special tokens: [<SINGLE VEHICLE WITH OBJECT>, <ANGLE IMPACTS\_RIGHT>, <OTHER>, <SIDESWIPES\_LEFT>, <FRONT END COLLISIONS>, <REAR END COLLISIONS>, <OVERTURN>, <ANIMAL COLLISIONS>, <PEDESTRIAN COLLISIONS>, <SIDESWIPES\_RIGHT>, <PEDALCYCLIST COLLISIONS>, <HEAD ON COLLISIONS>, <OFF ROAD>, <ANGLE IMPACTS\_LEFT>], each representing a specific crash type.

\paragraph{The Generation of Satellite Images}
HSIS provides the coordinates of crash locations in the Washington State Plane South coordinate system. This system uses the Washington coordinate system of 1983, South Zone, which is a Lambert conformal conic projection based on the GRS 80 spheroid. The standard parallels for this projection are located at north latitudes 45° 50' and 47° 20', where the scale is exact. The origin of this coordinate system is defined at the intersection of the meridian 120° 30' west of Greenwich and the parallel 45° 20' north latitude, with assigned coordinates: E = 500,000 meters and N = 0 meters~\footnote{https://business.wsdot.wa.gov/}.
To obtain the satellite images, we convert these coordinates into GPS coordinates (latitude and longitude). We then use the Google Maps API to request satellite images with a resolution of 512 $\times$ 512 pixels and a zoom level of 19.

\begin{figure}[h]
  \centering
  \includegraphics[width=0.99\textwidth]{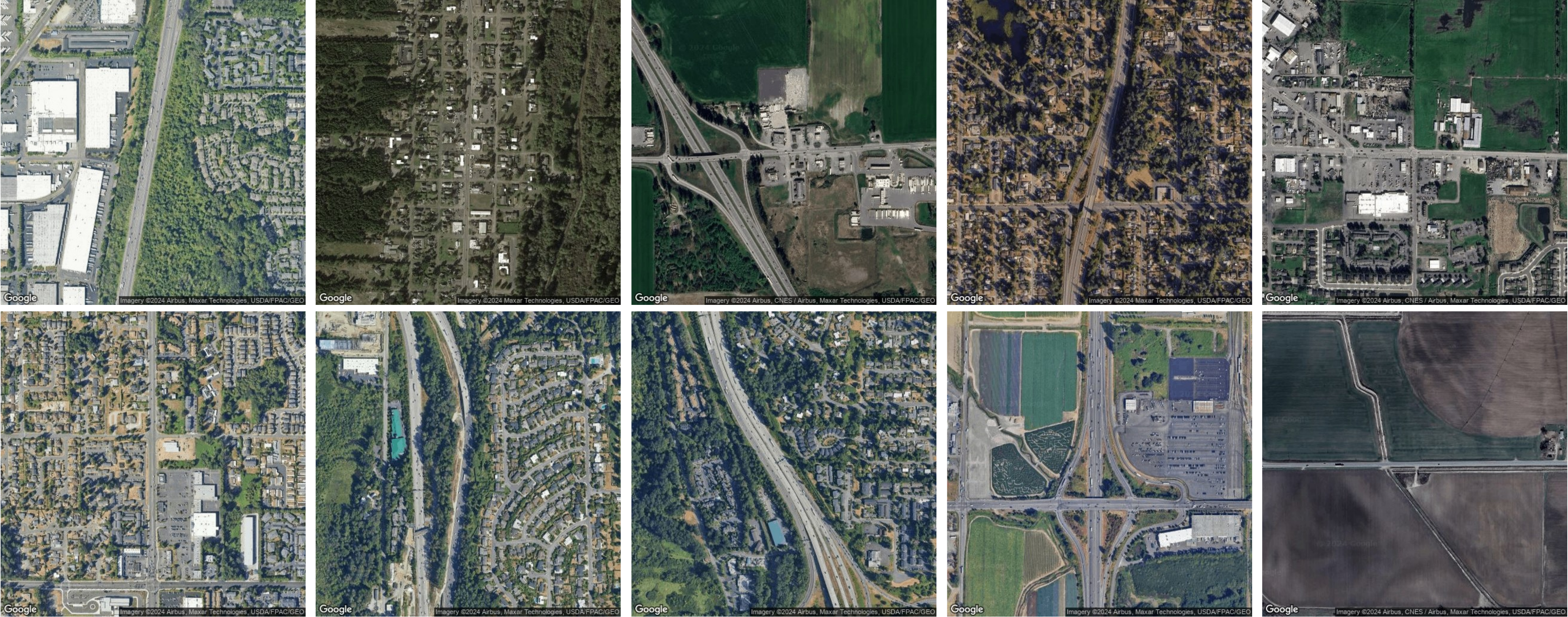}
\caption{\textbf{Satellite Images Generation} by querying Google Map service from HSIS datasets.}\label{fig:example_satellite}
\end{figure}

\paragraph{Explanation about What-if Analysis.} What-if analysis is a powerful technique used to understand the impact of changes in input variables on the output of a model. This method allows researchers and decision-makers to explore various scenarios by modifying input parameters and observing the subsequent changes in model predictions.
In practice, what-if analysis involves altering specific features in a dataset to evaluate how these changes affect the model's output. For instance, in a traffic crash prediction model, we might modify driver conditions to investigate how these factors influence the likelihood of an accident. This approach is instrumental in identifying key factors that significantly impact outcomes and helps in developing more robust and interpretable models.
Specifically, in Figure 4 of the main draft, we analyze the effects of three factors: "driving under or without alcohol," "icy or dry road conditions," and "within or outside a work zone." We examine 842 test examples distributed across January, June, and December.

Consider the "driving under or without alcohol" scenario as an example. Among the 842 test cases, there are 63 crashes involving alcohol and 779 cases without alcohol involvement. To perform the what-if analysis for the alcohol variable, we randomly select an additional 63 cases from the 779 non-alcohol cases, creating a set of 126 cases for the analysis, labeled as "alcohol (+100\%)." Additionally, we synthesize another 126 cases from 779 non-alcohol cases, formulating total 189 cases, and denote this set as "alcohol (+200\%)." Finally, we transform all 779 non-alcohol cases into alcohol-involved cases and conduct the analysis, labeled as "alcohol (all)."
Similarly, there are 

\begin{figure}[h]
  \centering
  \includegraphics[width=0.99\textwidth]{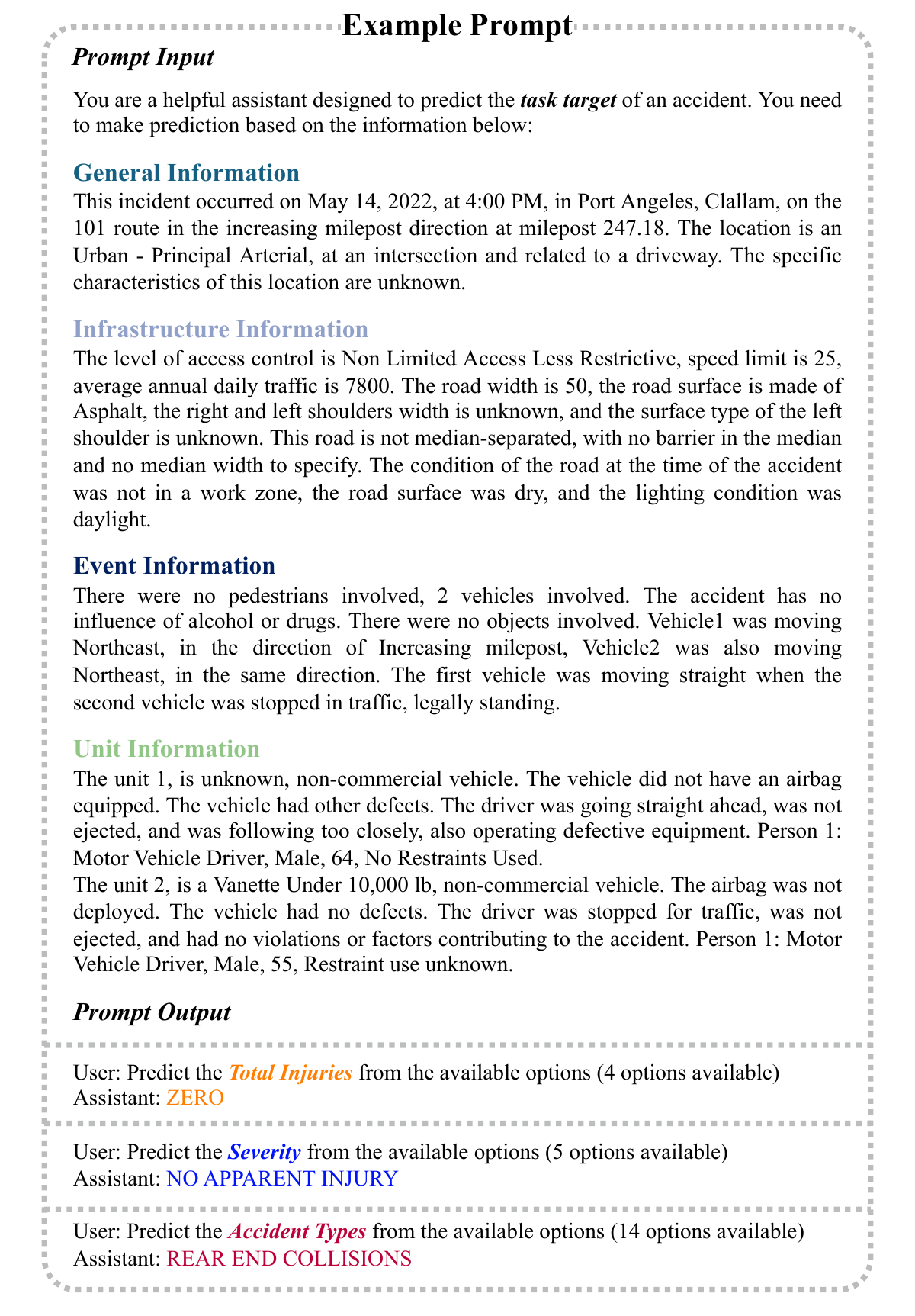}
\caption{\textbf{One of the prompt examples (1/5)} used in our CrashEvent to construct datasets for training.}\label{fig:example_prompt_1}
\end{figure}

\begin{figure}[h]
  \centering
  \includegraphics[width=0.99\textwidth]{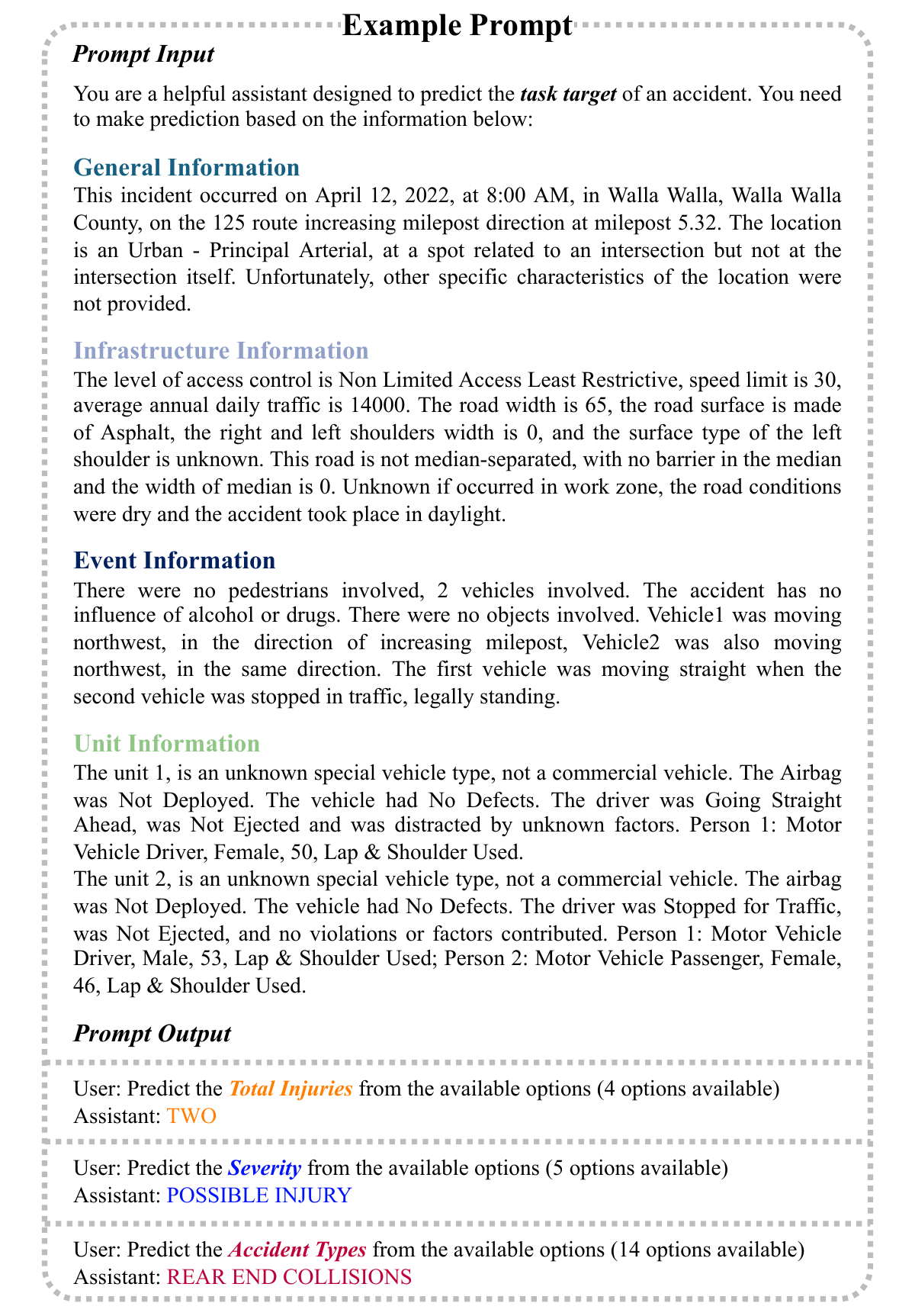}
\caption{\textbf{One of the prompt examples (2/5)} used in our CrashEvent to construct datasets for training.}\label{fig:example_prompt_2}
\end{figure}

\begin{figure}[h]
  \centering
  \includegraphics[width=0.99\textwidth]{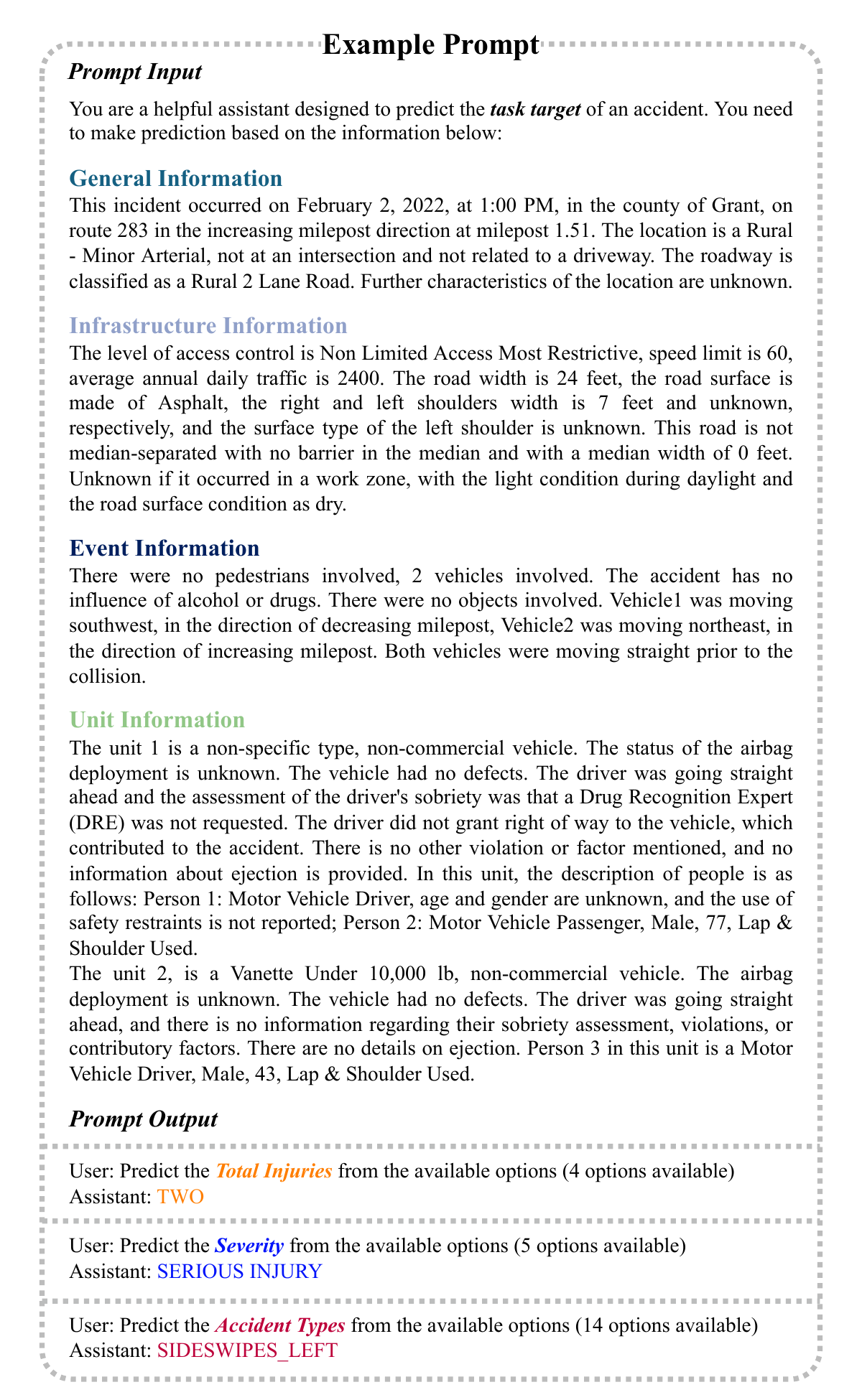}
\caption{\textbf{One of the prompt examples (3/5)} used in our CrashEvent to construct datasets for training.}\label{fig:example_prompt_3}
\end{figure}

\begin{figure}[h]
  \centering
  \includegraphics[width=0.99\textwidth]{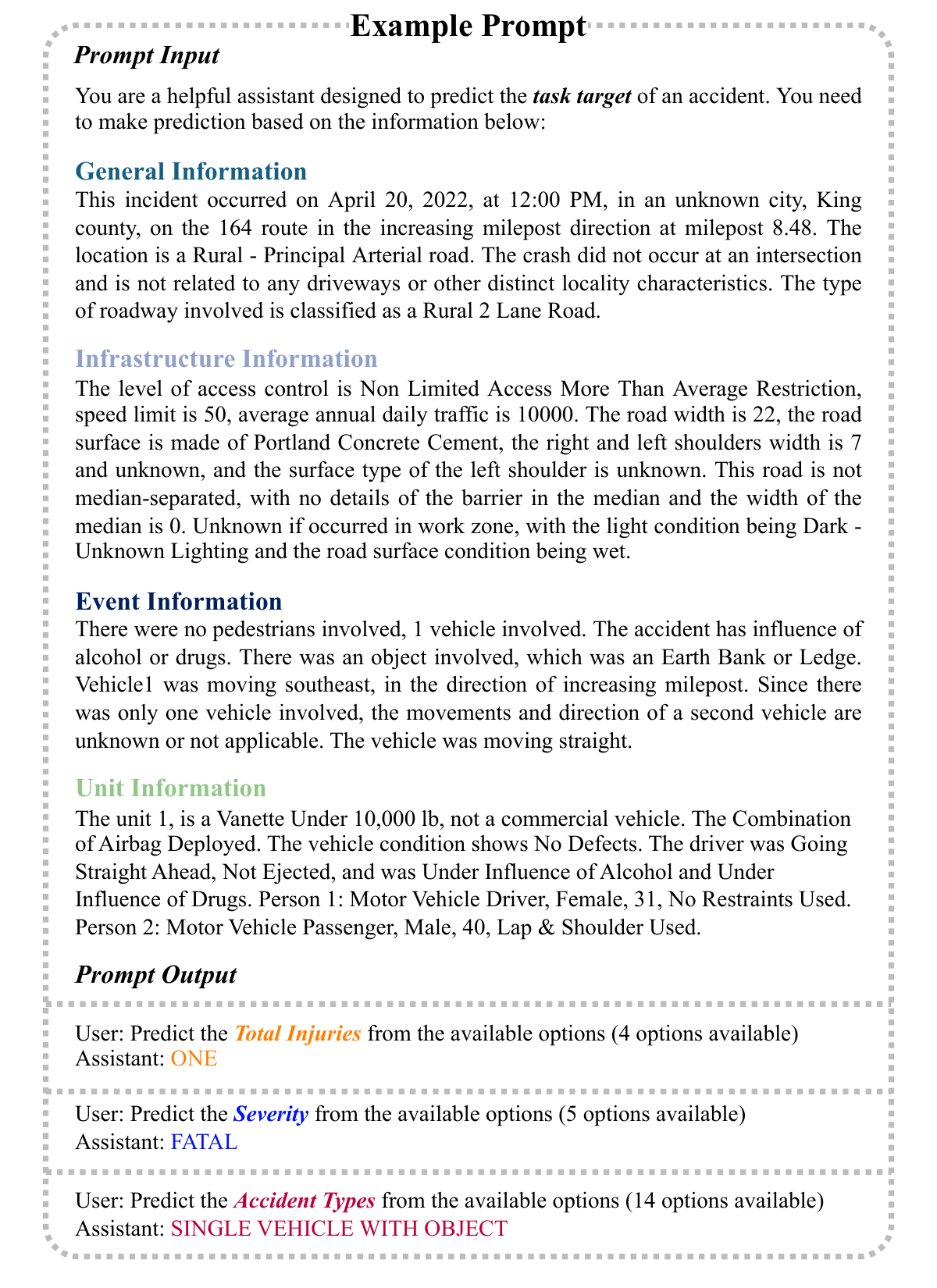}
\caption{\textbf{One of the prompt examples (4/5)} used in our CrashEvent to construct datasets for training.}\label{fig:example_prompt_4}
\end{figure}

\begin{figure}[h]
  \centering
  \includegraphics[width=0.99\textwidth]{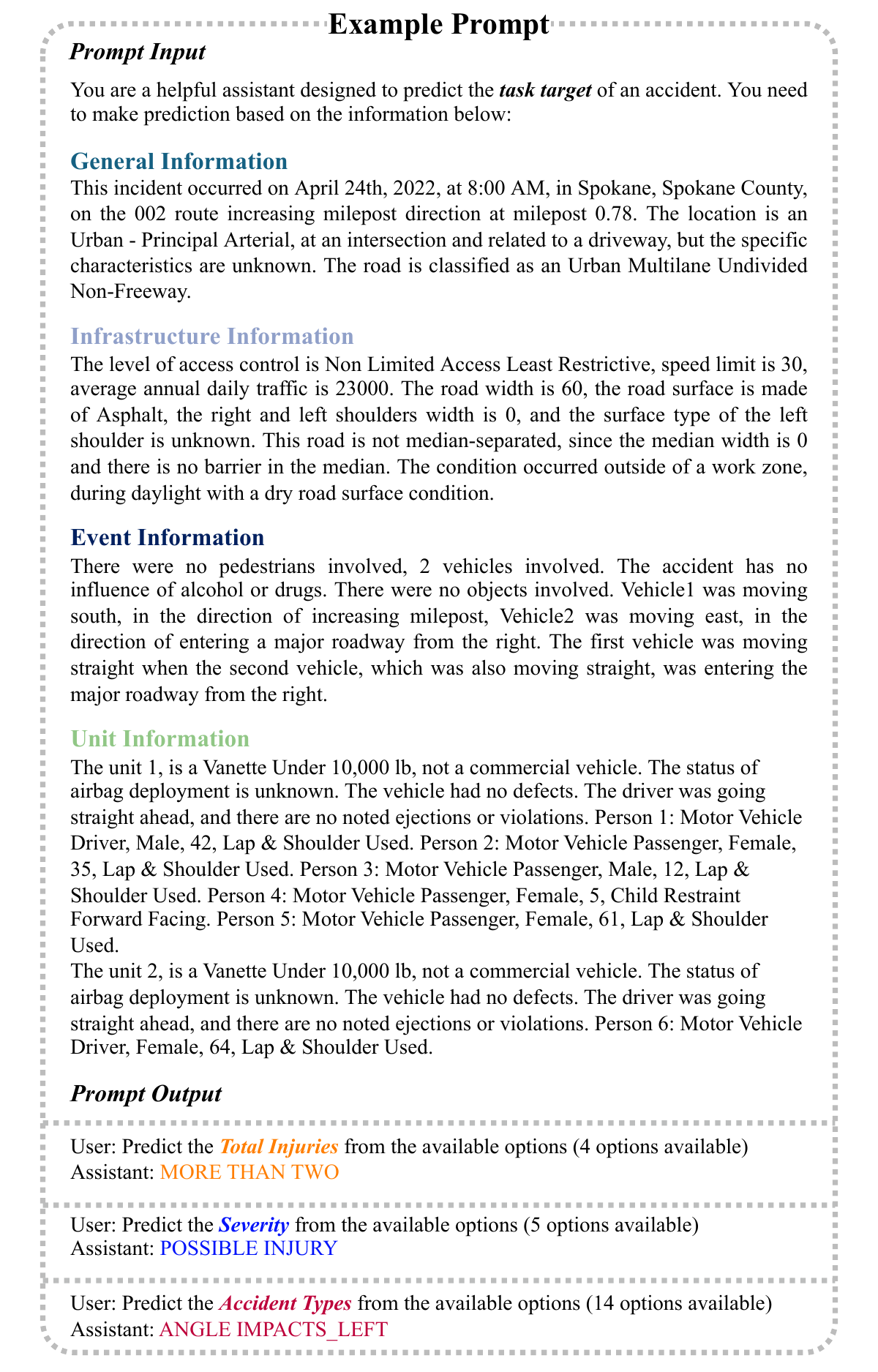}
\caption{\textbf{One of the prompt examples (5/5)} used in our CrashEvent to construct datasets for training.}\label{fig:example_prompt_5}
\end{figure}

\clearpage
\endgroup

{\small
\bibliographystyle{unsrt}
\bibliography{neurips_2024}
}

\clearpage

\end{document}